\documentclass[10pt]{article}
\usepackage[preprint]{tmlr}

\usepackage{amsmath,amsfonts,bm}

\def\eqref#1{equation~\ref{#1}}
\def\1{\bm{1}}

\DeclareMathAlphabet{\mathsfit}{\encodingdefault}{\sfdefault}{m}{sl}
\SetMathAlphabet{\mathsfit}{bold}{\encodingdefault}{\sfdefault}{bx}{n}

\usepackage{hyperref}
\usepackage{url}
\usepackage{amsmath,amssymb,amsthm}
\usepackage{booktabs}
\usepackage{graphicx}
\usepackage{xcolor}
\usepackage{enumitem}
\usepackage{float}
\usepackage{caption}
\usepackage{longtable}
\definecolor{linkblue}{HTML}{1F4E79}
\hypersetup{colorlinks=true,linkcolor=linkblue,citecolor=linkblue,urlcolor=linkblue}
\newtheorem{proposition}{Proposition}
\newtheorem{corollary}{Corollary}
\theoremstyle{definition}

\theoremstyle{remark}
\newtheorem*{remark}{Remark}

\newcommand{\cg}{CG-Mamba}
\newcommand{\pass}{\textsc{pass}}
\newcommand{\fail}{\textsc{fail}}
\newcommand{\ex}{exact}
\newcommand{\br}{broken}

\title{Certified, Not Explained: Exact Token Erasure\\in Gated Vision State Space Models}

\author{\name Sadia Parveen \\ \addr Department of Computer Science, COMSATS University Islamabad, Pakistan
\AND \name Dr. Tehseen Zia \\ \addr Department of Computer Science, COMSATS University Islamabad, Pakistan}

\def\month{MM}
\def\year{YYYY}

\begin{document}
\maketitle

% =====================================================================
\begin{abstract}
Vision models built on Mamba are often explained through their selective step size $\Delta$, on the assumption that a token the scan does not admit cannot influence the prediction. We test that in a concept bottleneck model for lung-nodule malignancy built on a single gated, bidirectional selective state-space block. The answer becomes an interpretability framework: a guarantee, a procedure that checks it, and a pre-registered audit of what it explains. For a fixed gate pattern, masking the input of the causal convolution of the block and giving closed tokens zero weight at readout makes the concept and task logits exactly invariant to the closed tokens. Gating $\Delta$ is neither necessary nor sufficient. On LIDC-IDRI, a path-by-path knockout over 15 cross-validated models agrees with this in every run: the largest logit change is exactly 0 whenever the conditions hold and between 4.1 and 6.7 when they do not. The gated model reaches a pooled out-of-fold AUROC of 0.8425 $\pm$ 0.0085, level with an ungated Mamba bottleneck (0.8428). The guarantee belongs to the masked model, not to the gate as an explanation. Post-hoc rankers, which inherit the guarantee on this model, order tokens better than the gate. Gate maps from a model trained on permuted labels are as close to the real maps as two real-label seeds are to each other. Under retraining, the tokens of the gate carry more task information than random tokens and, on shifted crops, more than central ones, but no more than the tokens chosen by hidden attention. We therefore read the gate as a certified support set produced by mask supervision. Splitting the Oracle Impurity Score into own-concept and cross-concept parts reveals differences the aggregate score misses. Partitioning the readout reduced leakage slightly under an earlier recipe, and the final model sits at, rather than below, the level implied by correlations among the true concepts.
\end{abstract}

% =====================================================================
\section{Introduction}
\label{sec:intro}

Concept bottleneck models (CBMs) put a layer of human-readable variables between the image and the label \citep{koh2020concept}. For pulmonary nodules on CT the natural variables already exist: radiologists rate spiculation, margin, lobulation and several other attributes when they read a scan, and the LIDC-IDRI database records those ratings for every nodule of 3\,mm or larger that a reader marked \citep{armato2011lidc}. A CBM trained on them can tell a clinician \emph{what} it believes about a nodule and lets the clinician correct a belief before the malignancy score is computed \citep{shen2019hscnn,lalonde2020xcaps}.

What a CBM does not say is \emph{where} in the image each belief came from. That question is normally handed to a saliency method, and saliency methods can look convincing while being insensitive to the model they claim to describe \citep{adebayo2018sanity}. A second, separate worry is leakage: concept representations may carry information about other concepts or about the label beyond the predicted concept values themselves \citep{mahinpei2021promises,margeloiu2021concept,havasi2022leakage}. The standard probe-based measure of this, the Oracle Impurity Score (OIS; \citealp{espinosazarlenga2023metrics}), mixes several effects together, as we show in Section~\ref{sec:res-leak}.

Selective state-space models \citep{gu2023mamba} and their vision versions \citep{zhu2024vim,liu2024vmamba} seem to offer a built-in answer to the \emph{where} question. Each token enters the recurrent state through an input-dependent step size $\Delta$, and a token with $\Delta=0$ does not update the state. This suggests that $\Delta$, or a gate placed on it, could serve as a faithful map of what the model used. We started this project from that idea, built the model around it, and found that it is half right. A hard token gate inside the block can indeed give an exact guarantee, but the guarantee comes from a different path than $\Delta$, and it says less about explanation quality than we had hoped.

Stated after the fact, the core result can look obvious: set the input of the convolution to zero and nothing downstream sees the token. It was not what we expected, and it is not how the step size is usually discussed. $\Delta$ is described as the selection mechanism of these models \citep{gu2023mamba}, so a gate on $\Delta$ is where one would naturally place a guarantee. Our analysis shows that such a gate is neither necessary nor sufficient, and the knockout checks this on trained models, where an algebraic argument can fail if a single path has been overlooked.

This paper reports both halves. Together, the guarantee, the knockout procedure that verifies it, and the pre-registered tests of what it does and does not explain form the interpretability framework we propose for vision state space models. Our contributions are:
\begin{enumerate}[label=(\roman*),leftmargin=*,itemsep=2pt]
\item \textbf{Sufficient conditions for exact erasure in a gated selective SSM block} (Section~\ref{sec:theory}). With the gate pattern held fixed, masking the convolution input makes the entire selective scan independent of closed-token content; together with zero readout weight on closed tokens, the concept and task logits become exactly invariant. Residual and $z$-branch gating are required only when blocks are stacked. The $\Delta$ gate is never required.
\item \textbf{Confirmation on trained models} (Section~\ref{sec:res-ko}). A ten-configuration knockout on 15 cross-validated models of the final architecture reproduces the predicted pattern in every run, at both the concept and the task logit. The knockout does not depend on our model: it can audit any architecture that claims erasure through a hard token mask.
\item \textbf{A pre-registered audit of what the guarantee buys} (Sections~\ref{sec:res-acc}--\ref{sec:res-maps}). The gated model costs no accuracy relative to an ungated Mamba CBM, but its gate maps fail a label-randomisation test, rank tokens worse than MambaLRP, hidden attention and gradient$\times$input, and beat a centre prior under retraining only after the crops are shifted off the nodule. All pass/fail criteria were written down before the corresponding experiment was run.
\item \textbf{A decomposition of OIS} (Section~\ref{sec:res-leak}) into own-concept and excess cross-concept predictability. It detects a small, dimension-controlled leakage reduction from partitioning the readout channels by concept under an earlier recipe, and shows that the final model sits at the level implied by concept correlations; OIS registers neither.
\end{enumerate}

\paragraph{What this paper does not claim.} Our crops have 64 tokens, so attention is cheap, and zeroing a gate does not skip computation. We make no efficiency or scaling claim for the SSM (Section~\ref{sec:discussion}). Nor is the guarantee unique to state-space models: any network that hard-masks its input tokens and excludes them at readout can obtain exact erasure. What is specific to the selective SSM is concrete, though. A token can reach the output of the block through five internal paths. The one that the Mamba literature points to, $\Delta$, turns out not to matter, and the one that does, the convolution input, is rarely discussed. Listing the paths is what makes the guarantee checkable rather than assumed, and the mask itself is computed inside the block.

% =====================================================================
\section{Related work}
\label{sec:related}

\paragraph{Concept bottlenecks and leakage.} Hard and soft CBMs \citep{koh2020concept} and concept embedding models \citep{espinosazarlenga2022cem} trade some accuracy for an intervenable concept layer. Soft concept outputs can encode information beyond the concept values themselves, which the label head can exploit \citep{mahinpei2021promises,margeloiu2021concept}; this inflates accuracy and makes interventions less reliable \citep{havasi2022leakage}. Recent work quantifies such leakage with information-theoretic measures, both between concepts and from concepts to the task \citep{makonnen2025measuring,parisini2025leakage}. \citet{espinosazarlenga2023metrics} introduced the OIS and NIS probe metrics; \citet{raman2024locality} showed that CBMs often ignore spatial and concept locality. On LIDC-IDRI, concept-based malignancy models range from hierarchical CNNs \citep{shen2019hscnn} and capsule networks \citep{lalonde2020xcaps} to recent bottlenecks distilled from CT foundation models or trained with weak priors \citep{dumaev2024concept,tushar2026distilling,ma2026less}. These use different nodule selections, label definitions and evaluation protocols, so we do not compare their numbers with ours. Concept bottlenecks are also being adopted for other clinical tasks, for example glioblastoma response assessment on longitudinal MRI \citep{tarek2026trace}.

\paragraph{Faithful by construction.} In NLP, select-then-predict models obtain faithful rationales by letting the predictor read only a selected subset of the input \citep{lei2016rationalizing,bastings2019interpretable,jain2020fresh}. \citet{bastings2019interpretable} relax the binary selection with the same hard-concrete distribution \citep{louizos2018l0} we use. Prototype networks \citep{chen2019protopnet} and B-cos networks \citep{bohle2022bcos} also build the explanation into the forward pass, in line with the argument that high-stakes models should be interpretable by design \citep{rudin2019stop}. ERASER \citep{deyoung2020eraser} made the comprehensiveness and sufficiency of such rationales measurable, and we borrow its sufficiency measure in Section~\ref{sec:res-faith}. In the vocabulary of \citet{jacovi2020faithfulness}, our guarantee is a faithfulness statement about what the model can depend on; it says nothing about plausibility. It is of the select-then-predict type, and what we add is the analysis of which internal paths of a selective SSM block a selection mask has to close.

\paragraph{Explaining Mamba.} MambaLRP \citep{jafari2024mambalrp} adapts layer-wise relevance propagation to selective SSMs so that relevance is conserved through the scan. \citet{ali2025hidden} rewrite the selective scan as a data-dependent attention matrix and derive rollout and attribution maps from it. Both are post hoc; we run both on our trained model and compare them with the gate (Section~\ref{sec:res-faith}). More recent methods target vision Mamba directly. X-VMamba \citep{mabrok2025xvmamba} scores image patches by how strongly they can steer the hidden state, EVM-Fusion \citep{yang2025evmfusion} builds maps of the step size $\Delta$ into a medical image classifier, and \citet{mohan2026interpreting} identify activation subspaces in Mamba models that can be used to interpret and steer them. Most of this work places token selection in $\Delta$ or in the state dynamics. We ask a different question: which of the block's paths a guarantee about ignored tokens actually depends on.

\paragraph{Evaluating attributions.} Deletion curves \citep{samek2017evaluating,petsiuk2018rise} ask how quickly the output moves when top-ranked inputs are removed. ROAR \citep{hooker2019roar} retrains after removal so that the model is never scored on inputs it has not seen. Model and data randomisation tests \citep{adebayo2018sanity} ask whether a map changes when what the model learned changes. We use all three, because each one catches a failure the others miss.

% =====================================================================
\section{Method}
\label{sec:model}

\begin{figure}[t]
\centering
% Figure 1 is the authors' own architecture drawing (draw.io). Export it as figures/fig_arch.pdf.
\IfFileExists{figures/fig_arch.pdf}{%
\includegraphics[width=\textwidth]{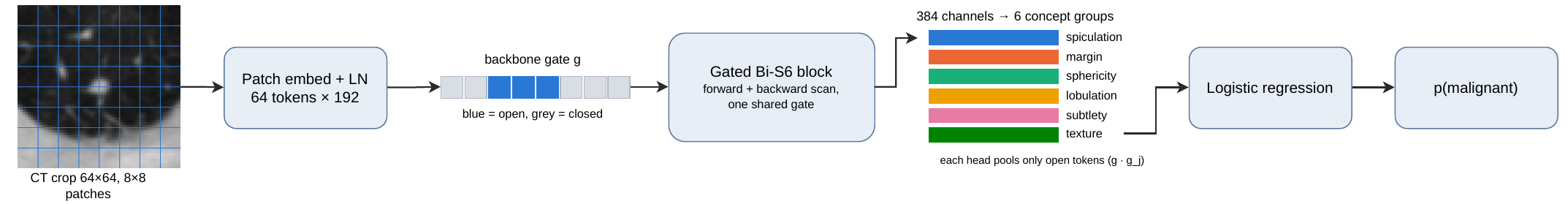}}{%
\fbox{\parbox[c][5.2cm][c]{0.96\textwidth}{\centering\small
Figure 1 placeholder: place the authors' architecture drawing at \texttt{figures/fig\_arch.pdf}.}}}
\caption{\textbf{\cg{}.} A $64{\times}64$ CT crop is cut into $64$ patch tokens. A backbone gate $g^b$, computed from each embedded token, closes tokens inside one bidirectional gated S6 block; the forward and the flipped backward scan share the same gate. The block's $384$ hidden channels are split into six groups of $64$, one per radiologist-rated concept. Each concept head pools its own group with weights $g^b_u g^j_u$, and the malignancy score is a logistic function of the six concept probabilities only.}
\label{fig:arch}
\end{figure}

Our approach has three parts. The first, described in this section, is a concept bottleneck model whose image encoder is a single gated, bidirectional selective state-space block. A learned gate decides, for every image patch, whether that patch may contribute to the concepts, and a closed patch is cut off at each point where it could otherwise enter the computation. The second part, Section~\ref{sec:theory}, states the conditions under which this cut-off is exact. The third, Section~\ref{sec:protocol}, describes how we test the guarantee and how much it explains. Below we first recall how a selective state-space block processes a sequence of tokens (Section~\ref{sec:background}), then show where the gate enters the block (Section~\ref{sec:block}), how the concepts are read out (Section~\ref{sec:cbm}) and how the model is trained (Section~\ref{sec:training}). Figure~\ref{fig:arch} gives an overview.

\subsection{Background: selective state-space blocks}
\label{sec:background}

A state-space layer reads a sequence of tokens in order and keeps a hidden state that summarises what it has seen so far. In Mamba \citep{gu2023mamba} the update depends on the current token through a step size $\Delta_u$. A large $\Delta_u$ lets the token rewrite much of the state, while $\Delta_u$ close to $0$ leaves the state almost unchanged; this input dependence is what is usually called \emph{selection}. Before the state update, each token passes through a short causal convolution that mixes it with its immediate predecessors. After the update, the output is multiplied by a second branch $z$ computed from the same token, and a residual copy of the input is added. Vision versions such as Vim \citep{zhu2024vim} apply this block to a sequence of image patches and scan it in both directions. What matters for this paper is that a token can reach the output through several of these routes, not only through $\Delta$.

\subsection{Gated selective state-space block}
\label{sec:block}

A crop is split into $N=64$ non-overlapping $8{\times}8$ patches. A strided convolution embeds each patch into $\mathbb{R}^{192}$ and a LayerNorm follows, giving tokens $x_1,\dots,x_N$. The backbone gate is a hard-concrete variable \citep{louizos2018l0} with temperature $2/3$ and stretch interval $(\gamma,\zeta)=(-0.1,1.1)$ whose location is a linear function of the token, $s_u = w^\top x_u + b$. During training $g_u$ is sampled. At evaluation it is deterministic,
\begin{equation}
g_u = \min\!\big(1, \max\big(0,\; 1.2\,\sigma(s_u) - 0.1\big)\big),
\label{eq:gate}
\end{equation}
which is a continuous value in $[0,1]$ that equals $0$ \emph{exactly} whenever $\sigma(s_u)\le 1/12$, that is, $s_u\le -\ln 11\approx-2.398$. We call such tokens \emph{closed} and write $S=\{u: g_u=0\}$. Tokens with $g_u>0$ are open, some of them only partly.

Given $g$, the forward-direction block computes
\begin{align}
(a_u, z_u) &= W_{\text{in}}\,\mathrm{LN}(x_u), \qquad c_u = g_u\, a_u, \label{eq:convin}\\
\tilde a_u &= \mathrm{SiLU}\Big(\textstyle\sum_{k=0}^{3} w_k \odot c_{u-k}\Big), \\
(\delta_u, B_u, C_u) &= W_x\,\tilde a_u, \qquad \Delta_u = g_u\,\mathrm{softplus}(W_\Delta \delta_u + b_\Delta), \label{eq:delta}\\
h_u &= \exp(\Delta_u A)\odot h_{u-1} + \Delta_u B_u\, \tilde a_u, \qquad y_u = C_u h_u + g_u\, D\odot \tilde a_u, \label{eq:scan}\\
\tilde y_u &= y_u \odot \mathrm{SiLU}(g_u z_u), \qquad o_u = W_{\text{out}}\, \tilde y_u + g_u\, x_u, \label{eq:out}
\end{align}
with $h_0=0$, a depthwise causal convolution of width $4$, state size $16$, $\Delta$-rank $12$ and inner width $384$. A closed gate therefore touches five paths: the convolution input (Eq.~\ref{eq:convin}), the step size (Eq.~\ref{eq:delta}), the $D$-skip (Eq.~\ref{eq:scan}), the multiplicative $z$ branch and the residual (Eq.~\ref{eq:out}). The backward direction applies the same equations, with separate weights, to the reversed sequence and the reversed gate pattern, and the two hidden sequences $\tilde y$ are averaged. Computing one gate and sharing it between directions matters: a token closed in one scan and open in the other would reach the readout through the open one.

\subsection{Concept heads and the readout partition}
\label{sec:cbm}

The averaged hidden states are layer-normalised and cut into $J=6$ contiguous channel groups $\tilde y^{(j)}_u\in\mathbb{R}^{64}$, one per concept: spiculation, margin, sphericity, lobulation, subtlety and texture. (Calcification and internal structure are rated on categorical rather than ordinal scales in LIDC-IDRI and are not used.) We refer to this split as the \emph{readout partition}: concept $j$'s head can only read channel group $j$. Each head has its own hard-concrete gate $g^j_u$, computed from $\tilde y^{(j)}_u$ by a linear map in the same way as Eq.~\ref{eq:gate}, and forms
\begin{equation}
\hat c_j = \sigma\!\Big(v_j^\top \frac{\sum_u w^j_u\, \tilde y^{(j)}_u}{\sum_u w^j_u + \epsilon} + d_j\Big), \qquad w^j_u = g^b_u\, g^j_u, \qquad \epsilon = 10^{-6}.
\label{eq:pool}
\end{equation}
The malignancy probability is a logistic regression on $(\hat c_1,\dots,\hat c_6)$ alone, so an expert can overwrite any $\hat c_j$ at test time.

\subsection{Training}
\label{sec:training}

Training has two stages. Nodule masks are available for a random 50\% of the training patients in each fold, the same patients for every seed; the other half receive no localisation signal.

\emph{Stage A} (150 epochs) trains the whole network with the backbone gate active. Each concept head pools its channel group with the backbone-gate weights $g^b$ only. The loss has four terms:
\begin{itemize}[leftmargin=*,itemsep=1pt,topsep=2pt]
\item a concept loss;
\item a hinge on backbone-gate logits above $-3$ (weight $3\times10^{-4}$);
\item a recall term, $1-\text{recall}$, that penalises closing tokens inside the nodule mask on masked images (weight $1$);
\item a budget term, $\mathrm{relu}(\bar g - 0.35)$, on each image's mean gate value (weight $5$).
\end{itemize}

\emph{Stage B} (100 epochs) freezes the patch embedding, the embedding norm and the backbone gate. It re-initialises the concept predictors, adds the concept gates and trains everything except the label head. Its loss has three terms:
\begin{itemize}[leftmargin=*,itemsep=1pt,topsep=2pt]
\item the concept loss;
\item a hinge on concept-gate logits over open tokens (weight $0.01$);
\item a localisation loss, $1-\text{Dice}$, between each concept's pooling weights $g^b g^j$ and the token-level nodule mask on masked images.
\end{itemize}

In both stages the non-concept terms are ramped in linearly over the first 20 epochs. Both stages use AdamW with learning rate $5\times10^{-4}$, weight decay $10^{-4}$ and batch size 16, a five-epoch linear warm-up followed by cosine decay to 2\%, and gradient clipping at 1. The forward and backward mixers are initialised from layer 0 of a pretrained Vim-Tiny model \citep{zhu2024vim}. After training, the label head is fitted as an L2-regularised logistic regression ($C=1$, class-balanced) on the concept probabilities that the model predicts for the fold's training images. The final model has $518{,}806$ parameters, and on held-out images $66$--$69\%$ of tokens are closed. Appendix~\ref{app:hparams} lists all settings.

% =====================================================================
\section{When is erasure exact?}
\label{sec:theory}

Fix a gate pattern $g$ and its closed set $S$. We want conditions under which the concept and task logits do not change when the content $\{x_s: s\in S\}$ of closed tokens changes. This section gives the three statements and what follows from them; the proofs are short and are collected in Appendix~\ref{app:proofs}. Section~\ref{sec:res-ko} then tests each statement on trained models.

\begin{proposition}[Masking the convolution input empties the scan of closed content]
\label{prop:conv}
For fixed $g$, the quantities $\tilde a_u$, $\Delta_u$, $B_u$, $C_u$, $h_u$ and $y_u$ do not depend on $\{x_s : s\in S\}$, for every position $u$, open or closed.
\end{proposition}
The proof (Appendix~\ref{app:proofs}) never uses the $\Delta$ gate or the $D$-skip gate. Once the convolution input is masked, a closed token contributes nothing to $\tilde a$, so there is nothing left for $\Delta$ to suppress.

\begin{corollary}[Exact erasure at the logits]
\label{cor:logits}
Assume convolution-input masking. For an open token $u$, $\tilde y_u$ depends on $x_u$ through $z_u$ but not on $x_S$. For a closed token, $\tilde y_u = y_u\odot\mathrm{SiLU}(g_u z_u)$ still depends on $x_u$ unless $z$ is gated, in which case $\tilde y_u = 0$ because $\mathrm{SiLU}(0)=0$. Hence the concept logits, and the task logit computed from them, are exactly invariant to $x_S$ if \textup{(i)} closed tokens receive zero pooling weight, as they do when $g^b_u$ is a factor of $w^j_u$ in Eq.~\ref{eq:pool}, \emph{or} \textup{(ii)} $z$ is gated at closed tokens.
\end{corollary}

The per-token LayerNorm before the heads does not change this: it maps a content-free vector to a content-free vector.

\begin{proposition}[Stacked blocks]
\label{prop:depth}
If $z$ and the residual are also gated, then $o_u = 0$ for every $u\in S$, independently of $x_u$ (the output projection $W_{\text{out}}$ has no bias). A following block receives content-free input at those positions even if its own gate re-opens them, so Proposition~\ref{prop:conv} and Corollary~\ref{cor:logits} apply block by block. With convolution-input masking alone, $o_u$ at closed positions still carries $x_u$ through the residual and through $z$, and a later block that re-opens those positions leaks it.
\end{proposition}

\begin{remark}[What the guarantee covers]
All three statements hold the gate pattern fixed. In deployment the gate is itself a function of the image, and changing closed tokens can change the gate. For an earlier model trained without mask supervision, replacing the closed tokens by air on the 213 validation images and recomputing the gate re-opened every one of them, and the concept logits changed by $0.445$ on average. The guarantee is therefore conditional: \emph{given this mask}, the prediction cannot depend on what the mask hides. It does not say that any content would produce the same mask, and it says nothing about whether the open tokens are clinically meaningful.
\end{remark}

% =====================================================================
\section{Experimental protocol}
\label{sec:protocol}

\paragraph{Data.} We use LIDC-IDRI \citep{armato2011lidc}; the copy available to us contains patients 0001--0600. Each nodule is the cluster of annotations that \texttt{pylidc} assigns to it, and clusters with more than four annotations (merged nodules) are excluded. A nodule is labelled malignant when its mean malignancy rating exceeds 3 and benign when it is below 3; nodules whose mean rating lies within $0.5$ of 3 are dropped. Volumes are resampled to $0.75$\,mm isotropic voxels and a $64{\times}64{\times}33$ cube is cut around each nodule centroid. The model sees the central axial slice, clipped to $[-1000, 400]$\,HU and scaled to $[0,1]$. Concept targets are the readers' mean ratings, rescaled to $(r-1)/4$; the localisation mask is the 50\% consensus of the readers' contours. The cross-validation pool holds 764 nodules from 348 patients, $33.2\%$ of them malignant.

\paragraph{Splits, and a test set we do not use.} All results come from one fixed patient-level, stratified 5-fold split of the pool, trained with 3 seeds (15 models per configuration); the seeds change initialisation and data order, not the folds. Held-out folds contain 133, 155, 157, 157 and 162 nodules. We report pooled out-of-fold (OOF) AUROC per seed and its mean and sample standard deviation over seeds. A separate test split of 183 nodules exists, but it was evaluated at three points during development, after which the pipeline changed. It can no longer give an unbiased estimate, and no number in this paper comes from it. The cross-validation pool also contains the former validation split on which some hyperparameters were chosen, so the cross-validated numbers are mildly optimistic for every model alike.

\paragraph{Pre-declared tests.} From the point at which the final architecture was fixed, every experiment had its pass criterion written down before it was run, for example ``exact in every run'' or ``95\% bootstrap upper bound below 0 and at least 4 of 5 folds negative''. Each table reports the verdict, and Appendix~\ref{app:predeclared} collects all of them, including the failures. Paired comparisons of AUROC use the DeLong test \citep{delong1988comparing,sun2014fast} on pooled OOF predictions. Bootstrap intervals over five folds are approximate, and the 15 runs share folds, so Wilcoxon tests over runs are approximate as well.

\paragraph{Baselines.} All baselines use the same crops, folds, seeds, augmentation (random $90^\circ$ rotations and flips), optimiser and concept targets. We train a small ResNet-style CNN \citep{he2016resnet} and a two-layer ViT \citep{dosovitskiy2021vit}, each as a CBM with the same sequential label head and as an unconstrained classifier, and the same Mamba block without gates, again in both forms. Parameter counts differ (CNN CBM $700{,}669$; ViT CBM $620{,}365$; Mamba $518{,}806$), and only \cg{} sees nodule masks; we return to both confounds where they matter.

\paragraph{Metrics.} \emph{Certificate}: with the gate pattern forced, the pixels of closed tokens are replaced by Gaussian noise with standard deviation 50 (on an intensity scale where the data lie in $[0,1]$) or by air ($-1000$\,HU). We record the maximum absolute change of the concept and task logits over all held-out images. \emph{Deletion}: tokens are replaced by air in the order given by a ranking, and the mean absolute change of the task logit is recorded after $k\in\{0,4,8,16,\dots,64\}$ removals. The area under this curve is divided by its maximum (``deletion $\div$max''); higher means the ranking finds output-relevant tokens sooner. Deletion scores are comparable within one model, not across models. \emph{Localisation}: the energy-based pointing game (EPG), i.e.\ the fraction of a map's mass that falls inside the consensus nodule mask.

% =====================================================================
\section{Results}
\label{sec:results}

\subsection{Which paths the guarantee needs}
\label{sec:res-ko}

Table~\ref{tab:ko} removes gated paths one at a time from the trained final model, forces the gate pattern, corrupts only the closed tokens, and measures the change at both the concept and the task logits on the held-out fold. Before each knockout we checked that the patched forward pass reproduces the unpatched model exactly.

\begin{table}[t]
\centering
\caption{\textbf{Leakage-path knockout} on the final model: 15 runs (5 folds $\times$ 3 seeds), held-out folds; AUROC is mean $\pm$ SD over the 15 runs. Corruption (Gaussian noise with SD 50, or air) touches closed tokens only, with the gate pattern forced. ``Exact'' means a maximum change of $0$ at both the concept and the task logits in that run. The AUROC column switches a path off at evaluation time in a model trained with all gates; it measures train/test mismatch and is not the accuracy that a model trained without that path would reach. Pre-declared K1 (full, no-$\Delta$ and conv-only exact in every run): \pass{}. K2 (no conv mask, and conv-only with $g^j$-only pooling, broken): \pass{}.}
\label{tab:ko}
\small
\begin{tabular}{lccccc}
\toprule
configuration & $\max|\Delta \hat c|$ & $\max|\Delta\,\text{task}|$ & exact runs & guarantee & AUROC (eval.\ time) \\
\midrule
all paths gated & 0 & 0 & 15/15 & \ex & $0.8409 \pm 0.0345$ \\
$\Delta$ gate removed & 0 & 0 & 15/15 & \ex & $0.8291 \pm 0.0366$ \\
$D$-skip gate removed & 0 & 0 & 15/15 & \ex & $0.8406 \pm 0.0344$ \\
$z$ gate removed & 0 & 0 & 15/15 & \ex & $0.8426 \pm 0.0338$ \\
residual gate removed & 0 & 0 & 15/15 & \ex & $0.8409 \pm 0.0345$ \\
conv input only & 0 & 0 & 15/15 & \ex & $0.8323 \pm 0.0370$ \\
all paths, $g^j$-only pooling & 0 & 0 & 15/15 & \ex & $0.8467 \pm 0.0359$ \\
\midrule
conv-input mask removed & 5.44 & 5.02 & 0/15 & \br & $0.8263 \pm 0.0513$ \\
conv input only, $g^j$-only pooling & 5.98 & 4.14 & 0/15 & \br & $0.8355 \pm 0.0338$ \\
no gating & 6.69 & 5.40 & 0/15 & \br & $0.7959 \pm 0.0443$ \\
\bottomrule
\end{tabular}
\end{table}

The table follows Section~\ref{sec:theory} line by line. Every configuration that masks the convolution input and keeps closed tokens out of the readout, either through $g^b$ in the pooling weight or through the $z$ gate, is exact in all 15 runs. The two configurations that violate a condition, and the ungated model, break it in all 15. Removing the $\Delta$ gate leaves the guarantee intact. An earlier model trained without mask supervision shows the same pattern on a single validation split (Appendix~\ref{app:ko-val}).

The AUROC column should not be read as the accuracy value of each gate. Switching off the $\Delta$ gate at evaluation time costs $0.012$ here and $0.043$ on the single validation split, but measuring what $\Delta$ gating contributes to accuracy would require retraining without it. Removing the $z$ gate slightly raises AUROC; $z$ gating matters only once blocks are stacked (Proposition~\ref{prop:depth}).

\paragraph{Depth.} With two untrained blocks and forced gates, Gaussian noise (SD 50) added to the embeddings of tokens closed in block 1 leaves the output unchanged under full gating, even when block 2 re-opens 30\% of them. With convolution-input masking alone and the same re-opening, the change at open positions reaches $2.05\times10^{2}$ ($13.1$ after pooling), as Proposition~\ref{prop:depth} predicts. A trained two-block model with full gating (15 runs) is exact on every held-out image.

\paragraph{Which gates depth needs.} Proposition~\ref{prop:depth} makes a sharper prediction: at depth, the $\Delta$ and $D$ gates are unnecessary, while $z$ and the residual are both required. We tested this directly in the same two-block setting with five random initialisations (pre-declared tests Z1 and Z2, both \pass{}). Gating only the convolution input, $z$ and the residual is exact. Leaving $z$ ungated lets re-opened positions leak, with a maximum change of $0.36$; leaving the residual ungated leaks far more, $2.25\times10^{2}$. The difference in size is expected, because the residual passes $x_u$ through unchanged, whereas $z$ reaches the output only multiplied by $y_u$.

\subsection{Accuracy, and the cost of the bottleneck}
\label{sec:res-acc}

\begin{figure}[t]
\centering
\includegraphics[width=\textwidth]{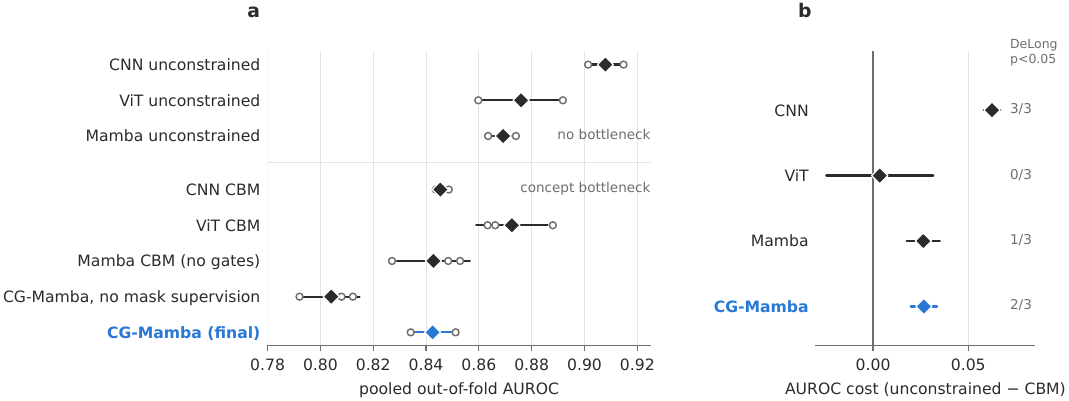}
\caption{\textbf{(a)} Pooled out-of-fold AUROC on 764 nodules (patient-level 5-fold CV). Open circles: the three seeds; diamonds and bars: mean $\pm$ sample SD over seeds. \textbf{(b)} Cost of the concept bottleneck on each backbone (unconstrained minus CBM, mean $\pm$ SD over seeds), with the number of seeds in which a paired DeLong test gives $p<0.05$. The CG-Mamba row uses the Mamba unconstrained model as reference.}
\label{fig:acc}
\end{figure}

\begin{table}[t]
\centering
\caption{\textbf{Task accuracy.} Pooled OOF AUROC per seed, mean $\pm$ sample SD over seeds, and concept Spearman correlation with the readers' mean ratings (mean $\pm$ SD over seeds).}
\label{tab:acc}
\small
\begin{tabular}{lccccc}
\toprule
model & seed 0 & seed 1 & seed 2 & mean $\pm$ SD & concept Spearman \\
\midrule
CNN, unconstrained & 0.9077 & 0.9148 & 0.9014 & $0.9079 \pm 0.0067$ & -- \\
ViT, unconstrained & 0.8598 & 0.8764 & 0.8918 & $0.8760 \pm 0.0160$ & -- \\
Mamba, unconstrained & 0.8740 & 0.8635 & 0.8701 & $0.8692 \pm 0.0053$ & -- \\
\midrule
CNN CBM & 0.8440 & 0.8486 & 0.8437 & $0.8454 \pm 0.0027$ & $0.5144 \pm 0.0168$ \\
ViT CBM & 0.8880 & 0.8633 & 0.8662 & $0.8725 \pm 0.0135$ & $0.5109 \pm 0.0295$ \\
Mamba CBM, no gates & 0.8529 & 0.8271 & 0.8484 & $0.8428 \pm 0.0138$ & $0.4105 \pm 0.0270$ \\
\cg{}, no mask supervision & 0.7921 & 0.8080 & 0.8123 & $0.8041 \pm 0.0107$ & $0.3543 \pm 0.0452$ \\
\textbf{\cg{} (final)} & 0.8422 & 0.8342 & 0.8512 & $\mathbf{0.8425 \pm 0.0085}$ & $0.4040 \pm 0.0257$ \\
\bottomrule
\end{tabular}
\end{table}

Figure~\ref{fig:acc} and Table~\ref{tab:acc} place the gated model among the baselines. It matches the ungated Mamba CBM (mean $\Delta=-0.0003$; DeLong $p>0.4$ in all three seeds) and the CNN CBM ($\Delta=-0.0029$, no seed significant), and trails the ViT CBM ($\Delta=-0.0300$, significant in 2 of 3 seeds). This comparison is not fully matched: \cg{} received mask supervision for half of the training patients, and no baseline did. A gate-free CBM cannot use that supervision, since the mask loss acts on gates. What the data support is therefore: \emph{with weak mask supervision, the gates cost nothing in task AUROC relative to an unsupervised, gate-free CBM.} Without mask supervision the gated model is clearly worse ($0.8041$).

Before the experiments we predicted that the bottleneck would cost at most one to two AUROC points on the Mamba backbone, and less than on the other backbones. The measured cost is $0.0267\pm0.0069$ (significant in 2/3 seeds). It is smaller than the CNN's ($0.0625$, 3/3) and larger than the ViT's ($0.0035$, 0/3), and outside the predicted band, so the prediction is partly confirmed and partly falsified. The gates add nothing on top of the bottleneck: the ungated Mamba CBM costs $0.0264$. Without a bottleneck, the Mamba classifier is below the CNN in every seed (mean $\Delta=-0.0387$, 3/3 significant) and level with the ViT ($-0.0068$, 0/3).

The larger weakness is concept quality. The Spearman correlation of \cg{} ($0.404$) is well below the CNN ($0.514$) and the ViT ($0.511$), and the ungated Mamba CBM ($0.411$) shares the deficit, so it is a property of this backbone rather than of the gates. A two-block gated model raised the correlation to $0.469$ (higher in all 15 runs; mean concept AUROC $0.744\to0.776$) while staying exact. Of our two pre-declared criteria, it met the concept one (mean concept AUROC $+0.032$, threshold $+0.02$) and missed the accuracy one (pooled AUROC $+0.008$ against a threshold of $+0.01$; no seed significant). It also remains below both baselines and has more parameters ($1{,}023{,}193$) than either. Depth closes part of the gap, at a capacity that is no longer matched.

\paragraph{Interventions.} Replacing predicted concepts by the readers' values in random order raises the mean AUROC of the final model towards $0.893$ at five replaced concepts, but the curve dips to $0.890$ at six and is monotone in only 7 of the 15 runs. In fold 1, predicted concepts beat the readers' values in all three seeds (for example $0.876$ with none replaced against $0.856$ with all six). We have not found the cause. Leakage confined to that fold and an unusual set of reader ratings would both produce it, and three seeds cannot separate the two. Intervention-aware training \citep{espinosazarlenga2023intcem} is designed for curves of this shape; we did not use it. None of our claims rests on intervention behaviour, so we report the anomaly and leave it open.

\subsection{How well does the gate rank evidence?}
\label{sec:res-faith}

\begin{figure}[t]
\centering
\includegraphics[width=\textwidth]{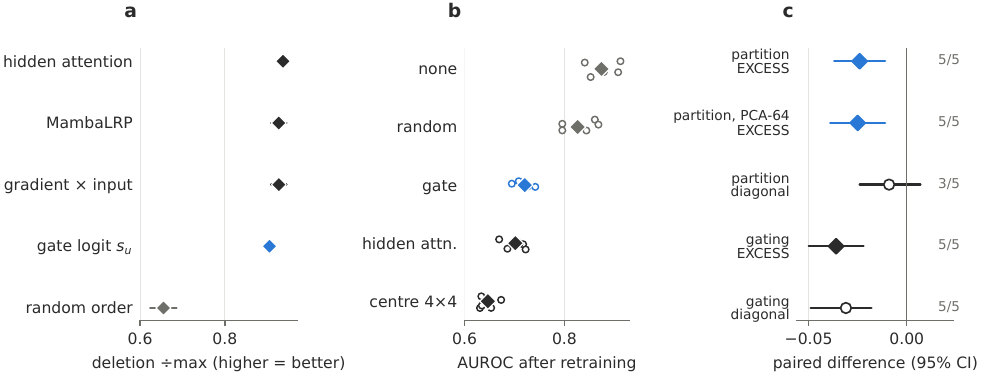}
\caption{\textbf{(a)} Deletion $\div$max on the final model, 15 runs (mean $\pm$ SD). \textbf{(b)} ROAR-lite on nodule-centred crops (shifted crops: Appendix~\ref{app:roar}): the top 16 of 64 tokens under each ranking are set to air in training and held-out images and an unconstrained Mamba classifier is retrained (seed 0). Open circles: the five folds; diamonds: pooled OOF AUROC. Lower means more task information was removed. \textbf{(c)} Leakage contrasts of Section~\ref{sec:res-leak}: paired fold-level differences with approximate 95\% bootstrap intervals and the number of folds below zero. Filled markers: EXCESS; open markers: diagonal.}
\label{fig:faith}
\end{figure}

\begin{table}[t]
\centering
\caption{\textbf{Deletion faithfulness} (deletion $\div$max on held-out folds, 15 runs; higher is better). Rows in one block share a random baseline; compare within blocks only. Sufficiency \citep{deyoung2020eraser}: largest change of the concept logits, averaged over images, after setting everything outside the method's top-22 tokens to air (CNN, ViT); for \cg{} the value $0$ is the conditional guarantee, a different quantity. Grad-CAM uses the predicted-class target with random tie-breaking.}
\label{tab:faith}
\small
\begin{tabular}{llccc}
\toprule
model & ranking & deletion $\div$max & random & sufficiency (method / random) \\
\midrule
CNN CBM & Grad-CAM & $0.7000 \pm 0.0898$ & $0.6833 \pm 0.1102$ & -- \\
CNN CBM & gradient$\times$input & $0.7847 \pm 0.0834$ & $0.6833 \pm 0.1102$ & $2.65$ / $4.49$ \\
\midrule
ViT CBM & gradient$\times$input & $0.8359 \pm 0.1041$ & $0.7933 \pm 0.0935$ & $2.48$ / $5.20$ \\
\midrule
\cg{} & gate logit $s_u$ & $0.9050 \pm 0.0135$ & $0.6553 \pm 0.0311$ & 0 (guarantee) \\
\cg{} & MambaLRP & $0.9273 \pm 0.0179$ & $0.6553 \pm 0.0311$ & -- \\
\cg{} & gradient$\times$input & $0.9273 \pm 0.0180$ & $0.6553 \pm 0.0311$ & -- \\
\cg{} & hidden attention & $\mathbf{0.9370 \pm 0.0109}$ & $0.6553 \pm 0.0311$ & -- \\
\bottomrule
\end{tabular}
\end{table}

\paragraph{Deletion.} Ranked by its own gate logit, \cg{} loses output much faster than under random deletion ($0.905$ against $0.655$, all 15 runs; Table~\ref{tab:faith}, Figure~\ref{fig:faith}a). Three post-hoc rankers on the same model do better: MambaLRP \citep{jafari2024mambalrp} ($0.927$; the gate wins in 3 of 15 runs, Wilcoxon $p=0.002$), gradient$\times$input \citep{shrikumar2016gradinput} ($0.927$) and the hidden-attention attribution of \citet{ali2025hidden} ($0.937$; the gate wins in 0 of 15, $p=0.0001$). Pre-declared tests M1 and M2 (MambaLRP and hidden attention beat random) pass; M3 and M4 (the gate beats each of them) fail. Our MambaLRP implementation reproduces the model's forward pass to $1.5\times10^{-5}$, gives closed tokens exactly zero relevance, and conserves relevance with a correlation of $0.883\pm0.022$ to the logit. It is no better than plain gradient$\times$input here (8 of 15 runs, $p=0.89$), plausibly because a single block leaves little room for the gradient shattering that LRP rules correct. The deletion metric uses logit change, which may favour gradient-based rankers.

On the CNN CBM, our first Grad-CAM \citep{selvaraju2017gradcam} implementation scored below random. The cause was index-ordered ties among zeroed tokens and a fixed positive-class target. After both were corrected, Grad-CAM sits at the random level ($0.700$ against $0.683$ on the final baseline models; $0.695$ against $0.697$ on the original ones, better than random in 5 of 15 runs, $p=0.85$).

\paragraph{ROAR-lite.} Deletion scores a model on inputs it never trained on. To rule this out, we removed the top 25\% of tokens under each ranking from both training and held-out images and retrained an unconstrained Mamba classifier from the same initialisation (Figure~\ref{fig:faith}b; per-fold values in Appendix~\ref{app:roar}). Removing the gate's tokens lowered pooled AUROC from $0.874$ to $0.721$, against $0.827$ for random removal (all five folds, DeLong $p<0.0001$; R1 \pass{}). Hidden attention removed a little more ($0.702$; R2 \pass{}; difference from the gate $p=0.157$). Removing simply the central $4{\times}4$ tokens of the nodule-centred crop removed more than either ($0.647$; worse than the gate in all five folds, $p<0.0001$), so R3, the gate beating the centre prior, fails. This criterion was poorly specified: the crops are centred on the nodule, so the central $4{\times}4$ tokens are largely the nodule itself, and removing them is close to removing the target. Section~\ref{sec:res-maps} makes the same point for EPG.

\paragraph{ROAR on shifted crops.} We therefore repeated the experiment with every crop shifted by 24\,px (three tokens) in one of four fixed directions, so that the central block no longer covers the nodule; the test was pre-declared before the run. Rankings were recomputed on the shifted images, and a new reference was retrained on shifted crops without removal (Appendix~\ref{app:roar}). The central $4{\times}4$ tokens now hold $10\%$ of the nodule mask, against $69\%$ for the gate's top 16. Removing them does not lower AUROC; it raises it, from $0.8516$ with no removal to $0.8859$, and it does so in all five folds (DeLong $p<0.001$). The most likely reason is regularisation: once the nodule has moved, the central block holds mostly parenchyma that carries little information about malignancy, and removing it leaves the classifier less to overfit. An artefact of the shift itself cannot be excluded. S1 and S2 compare removal conditions with each other, so this effect does not bear on them. Removing the gate's tokens, in contrast, lowers pooled AUROC to $0.6655$ (Table~\ref{tab:roar}): below the centre prior in all five folds (DeLong $p<10^{-10}$; S1 \pass{}) and below random removal ($0.8226$; all five folds, $p<10^{-11}$; S2 \pass{}). Once the centre prior no longer coincides with the nodule, the gate removes more task information than the prior. It still does not beat hidden attention ($0.6426$; difference $p=0.12$, with 81\% of the removed tokens shared), so this result says that the gate's open tokens carry the task information, not that the gate ranks them better than other Mamba attributions. Both ROAR experiments use one seed, one removal fraction ($25\%$) and a retrained evaluator that is not \cg{} itself; the shifted run also uses one shift per image.

\begin{table}[t]
\centering
\caption{\textbf{ROAR summary.} Pooled out-of-fold AUROC of an unconstrained Mamba classifier retrained after the top 16 of 64 tokens under each ranking are set to air (seed 0, five folds; lower means more task information removed). Nodule coverage: mean fraction of the consensus nodule mask inside the removed set, computed for the shifted crops. Per-fold values are in Appendix~\ref{app:roar}.}
\label{tab:roar}
\small
\begin{tabular}{lccc}
\toprule
tokens removed & centred crops & shifted crops (24\,px) & nodule coverage (shifted) \\
\midrule
none & 0.8740 & 0.8516 & -- \\
random & 0.8265 & 0.8226 & 0.25 \\
gate & 0.7206 & 0.6655 & 0.69 \\
hidden attention & 0.7017 & 0.6426 & 0.69 \\
centre $4{\times}4$ & 0.6466 & 0.8859 & 0.10 \\
\midrule
pre-declared & R1, R2 \pass{}; R3 \fail{} & S1, S2 \pass{} & \\
\bottomrule
\end{tabular}
\end{table}

\paragraph{What this means.} The gate is not a better ranking of evidence than post-hoc attribution, and it beats a simple spatial prior only when the cropping procedure does not place the nodule where the prior looks. Its distinctive property is the guarantee. On this model the post-hoc methods inherit that guarantee, because they assign zero relevance to closed tokens, which suggests using the two together: the gate fixes a support set that provably bounds what the prediction can use, and a post-hoc method ranks tokens inside it.

\subsection{What the gate maps reflect}
\label{sec:res-maps}

\begin{figure}[t]
\centering
\includegraphics[width=\textwidth]{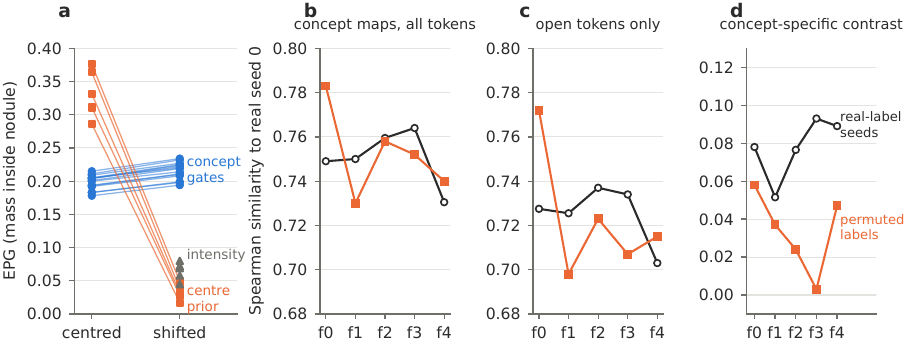}
\caption{\textbf{(a)} EPG of the final model's concept pooling weights on nodule-centred crops and on crops shifted by 16\,px (two tokens) in four directions; one line per run (15 runs for the gates, the five folds for the centre prior; triangles: intensity prior). \textbf{(b--d)} Spearman similarity of concept gate maps to those of the real-label seed-0 model, per fold. Black: other real-label seeds (the noise floor). Orange: a model retrained with concept and task labels permuted within each training fold (masks kept real). (b) all 64 tokens; (c) tokens open in both models; (d) after subtracting the component shared by the six concepts.}
\label{fig:maps}
\end{figure}

\paragraph{Localisation beyond trivial priors.} On centred crops a centre prior has a higher EPG than the gate ($0.334$ against $0.198$). That prior encodes the cropping procedure, not the image, so we shifted every crop by two tokens (16\,px). The ROAR experiment in Section~\ref{sec:res-faith} uses three tokens instead, because its centre set is the central $4{\times}4$ block rather than the central $2{\times}2$. The centre prior then falls to $0.033$, close to a uniform map ($0.030$), while the gate keeps $0.216$ and moves with the nodule (shift-follow $0.888\pm0.012$; Figure~\ref{fig:maps}a). On shifted crops the gate beats the centre prior and an intensity prior ($0.065$) in all 15 runs (P1, P2 \pass{}). The small rise from $0.198$ to $0.216$ probably comes partly from the zero-padded border and is not an improvement. Localisation depends on supervision: without masks, the gates of the earlier model reach an EPG of only $0.086$, which falsifies our pre-registered expectation that localisation would emerge without it.

\paragraph{Label randomisation.} We retrained the final recipe with concept and task labels permuted inside each training fold, keeping the masks real (seed 0, five folds). Task AUROC fell to $0.480\pm0.152$ (below $0.5$ in four of five folds; exact sign-flip permutation test against $0.5$, $p=0.69$) and mean concept AUROC to $0.501$. The predictions depend on the labels, and the guarantee stays exact, as a structural property must. The maps tell a different story (Figure~\ref{fig:maps}b--d). The permuted-label model's backbone gate is as similar to the real model's ($0.993\pm0.002$) as two real-label seeds are to each other ($0.992\pm0.001$), and it localises at least as well (EPG $0.207$). Its concept maps are as close to the real model's ($0.753\pm0.020$) as the seed-to-seed floor ($0.750\pm0.013$; below the floor in only 3 of 5 folds), so N1 fails. Restricting the comparison to tokens open in both models changes nothing ($0.723$ against $0.725$). Once the part shared by all concepts is removed, the concept-specific remainder is only weakly reproducible even between two real-label seeds ($0.078\pm0.016$; N3 fails). It is lower for the permuted model in all five folds ($0.034\pm0.021$), a consistent but small residual dependence on the labels that falls short of our pre-declared margin (N2 fails). That criterion bundled direction and size. The direction alone is consistent (one-sided exact sign test over the five folds, $p=0.031$), but a gap of about $0.04$, on a quantity that is only $0.08$ between two real-label seeds, is too small to call the concept-specific maps label-driven.

\paragraph{What this means.} The gates are a mask-supervised localisation of the nodule, shared across concepts. They carry no stable concept-specific pattern, and we do not present per-concept gate maps as explanations of concepts. The guarantee tells a reader which tokens the prediction \emph{can} depend on; it does not tell them why the model calls a nodule malignant.

\subsection{Leakage: what OIS misses}
\label{sec:res-leak}

\paragraph{A decomposition.} For each concept group $j$ we train a small probe (an MLP with 32 hidden units) on the pooled group-$j$ features to predict each binarised concept $k$, with thresholds chosen on fold-training data. This gives a $J\times J$ matrix $M$ of held-out probe AUROCs. The same probes trained on ground-truth concepts give an oracle matrix $M^{o}$. OIS is $\frac{2}{J}\lVert M-M^{o}\rVert_F$ \citep{espinosazarlenga2023metrics}. We separate four quantities: the mean diagonal (how well each concept is predicted from its own group), the mean off-diagonal (cross-concept predictability), $\mathrm{EXCESS}=\overline{\mathrm{off}}-\overline{\mathrm{off}}^{\,o}$ (cross-concept information beyond what correlations among the true concepts imply), and a diagonal-normalised version,
\begin{equation}
\mathrm{NEX} = \frac{\overline{\mathrm{off}}-0.5}{\overline{\mathrm{diag}}-0.5} - \frac{\overline{\mathrm{off}}^{\,o}-0.5}{\overline{\mathrm{diag}}^{\,o}-0.5}.
\end{equation}
In the five models of Table~\ref{tab:leak}, between 54\% and 60\% of $\mathrm{OIS}^2$ comes from the diagonal. OIS is largely measuring how weakly concepts are predicted from their own features, not how much they leak.

\begin{table}[t]
\centering
\caption{\textbf{Leakage decomposition} on held-out folds, 15 runs per model. Oracle off-diagonal $=0.680$; $\pm$ is the SD over runs. The first four models share the final architecture but use an earlier Stage-B recipe without mask supervision; the last row is the final model. ``Partition'': each concept head reads its own 64-channel group. For a dimension-matched comparison the unpartitioned 384-dimensional features are also reduced to 64 dimensions by PCA fitted on fold-training data ($86.8\%$ of variance retained). Paired contrasts are shown in Figure~\ref{fig:faith}c.}
\label{tab:leak}
\small
\setlength{\tabcolsep}{3.2pt}
\begin{tabular}{lccccccc}
\toprule
model & partition & gates & dim & OIS & diagonal & EXCESS & NEX \\
\midrule
partitioned, gated & yes & yes & 64 & $0.349 \pm 0.030$ & $0.677 \pm 0.024$ & $-0.030 \pm 0.027$ & $0.490 \pm 0.069$ \\
unpartitioned, gated & no & yes & 384 & $0.347 \pm 0.031$ & $0.686 \pm 0.020$ & $-0.006 \pm 0.026$ & $0.575 \pm 0.054$ \\
unpartitioned, gated, PCA-64 & no & yes & 64 & $0.347 \pm 0.028$ & $0.686 \pm 0.020$ & $-0.005 \pm 0.027$ & $0.578 \pm 0.072$ \\
partitioned, no gates & yes & no & 64 & $0.335 \pm 0.021$ & $0.708 \pm 0.025$ & $+0.006 \pm 0.022$ & $0.535 \pm 0.074$ \\
\midrule
final model & yes & yes & 64 & $0.344 \pm 0.019$ & $0.694 \pm 0.021$ & $+0.005 \pm 0.021$ & $0.597 \pm 0.094$ \\
\bottomrule
\end{tabular}
\end{table}

\paragraph{Partitioning.} Partitioning the readout channels lowers EXCESS by $0.024$ (95\% CI $[-0.037,-0.011]$, all five folds) and NEX by $0.085$ (all five folds), with no significant cost on the diagonal ($-0.009$, interval includes 0). The effect survives the dimension-matched comparison ($-0.025$, all five folds), so it is not an artefact of probe capacity. All four pre-declared criteria pass. OIS does not register the difference ($0.349$ against $0.347$).

\paragraph{Gating.} Comparing partitioned models with and without gates, gating also lowers EXCESS ($-0.036$, all five folds), but only together with weaker own-concept features (diagonal $-0.031$, all five folds); after normalisation the effect is inconsistent (3 of 5 folds). Here OIS even ranks the two models in the wrong direction ($0.349$ against $0.335$).

\paragraph{The final model.} The contrasts above use the earlier recipe. We repeated the decomposition on the 15 final models with the same probes, oracle and thresholds. EXCESS is $+0.005$ (fold-level 95\% CI $[-0.012,+0.021]$, negative in only 2 of 5 folds), so the pre-declared check that the final model stays below the oracle level (F1) fails. The final model does not leak beyond what correlations among the true concepts imply, but the suppression below that level which the partition produced under the earlier recipe is no longer present. The architecture is the same; the recipe is not. Mask-supervised training raised cross-concept predictability (EXCESS $+0.034$ relative to the earlier partitioned models, CI $[+0.025,+0.042]$, all five folds) more than own-concept predictability ($+0.017$). OIS again changes little ($0.344$ against $0.349$). Testing the partition under the final recipe would require unpartitioned models trained with it, which we have not run.

\paragraph{Limits.} The partition effect is small, about $0.025$ AUROC units. The unpartitioned model already sits at the oracle level ($\mathrm{EXCESS}\approx-0.006$), so it does not leak beyond what concept correlations imply; partitioning pushed cross-concept information \emph{below} that level, under the earlier recipe only. PCA to 64 dimensions is an imperfect dimension control.

% =====================================================================
\section{Discussion and limitations}
\label{sec:discussion}

\paragraph{Why a selective SSM?} The original motivation was scaling: selective SSMs are linear in sequence length, which should matter for volumetric or high-resolution images. Nothing here supports that. Our crops have 64 tokens, and gating zeroes values but does not skip computation, so all variants cost the same ($0.064$\,GFLOPs). A single 3D attempt ($64{\times}64{\times}32$ voxels, 512 tokens, one seed, one block, hyperparameters carried over from 2D) underfitted markedly. On the validation split the ungated Mamba CBM reached an AUROC of $0.719$ in 3D against $0.899$ in 2D, and the gated 3D model reached $0.812$ with a concept Spearman correlation of $0.158$. That is a statement about our compute and tuning budget, not evidence that 3D cannot work. This agrees with \citet{yu2025mambaout}, who argue that the SSM token mixer is not needed for image classification, where sequences are short and nothing has to be causal. What the selective SSM did offer is a precise account of which internal paths a token mask has to close, and a place in the block from which the mask can be read.

\paragraph{The guarantee is narrow on purpose.} It holds for a fixed gate pattern, does not cover changes to the gate, does not make the open set clinically meaningful, and is available to any architecture that hard-masks tokens and excludes them at readout. Its value is that it is checkable: a reader can verify, for any image, that nothing outside the support set can move the output. The checking procedure travels further than the model. Forcing the mask, corrupting what it hides and switching internal paths off one at a time works for any hard-masking architecture. Mamba-2 \citep{dao2024transformers} also passes a short causal convolution before its scan, so we expect a similar analysis to hold there, but we have not run it.

\paragraph{Data and statistics.} One public dataset; one axial slice per nodule, although several LIDC-IDRI concepts are defined on the volume; only patients 0001--0600; 764 nodules; no external validation. Because the test split was consulted during development, all evidence is cross-validated, and the pool includes the former validation split. The next step is an external estimate, on LIDC-IDRI patients 0601--1012, which were not available to us, or on an independent cohort. Several analyses (label randomisation, both ROAR experiments) use one seed.

\paragraph{Comparisons.} Backbones differ in parameter count, only the gated model saw nodule masks, and published LIDC-IDRI concept models with foundation-model encoders \citep{tushar2026distilling,ma2026less} use different cohorts and protocols.

% =====================================================================
\section{Conclusion}

A gated selective state-space concept bottleneck can carry an exact, checkable guarantee: for a given token mask, the concepts and the diagnosis cannot depend on what the mask hides. We derived sufficient conditions for that guarantee and confirmed them path by path on trained models, and found that masking the convolution input, not gating the step size $\Delta$, is what makes it hold. The same experiments mark its limits. The masks come from localisation supervision rather than from the labels, and they rank evidence no better than post-hoc methods; they beat a simple centre prior only once the crop no longer places the nodule at the centre. Separately, splitting the standard impurity score into its parts exposed differences that the aggregate score hides. Partitioning the concept readout reduced leakage slightly under an earlier recipe, and the final model sits at, not below, the level implied by correlations among the true concepts. Reporting the failed pre-declared tests next to the passing ones was, for us, the most informative part of the study.

% =====================================================================
\section*{Broader impact statement}
This work concerns interpretability for a diagnostic task. A certified support set could be mistaken for a clinically validated explanation; we have tried to say plainly that it is not, and that the gate maps reflect localisation supervision rather than evidence for malignancy. The model is a research prototype trained on one public dataset without external validation and must not be used for clinical decisions.

\section*{Reproducibility statement}
The code was extracted from the development notebook into a package and checked against it: 13 of 13 checks reproduce the notebook with a maximum difference of $0$, covering the data state, the forward passes of all 15 final models, AUROC, deletion scores, MambaLRP, hidden attention and the CNN and ViT baselines. The three experiments added in revision (ROAR on shifted crops, the depth-two path test and the leakage decomposition of the final model) are not covered by these checks. They were run as notebook cells on the same data state and checkpoints; the shifted-crop ROAR uses the package's explanation and metric modules, and the other two reuse notebook code that is identical to the code checked above. The code of the package and of the three revision experiments is publicly at \url{https://github.com/SadiaSatti095/CG-Mamba}. Fold assignments, checkpoints and every result file used in this paper will be released with the code. All settings are listed in Appendix~\ref{app:hparams}.

\bibliography{references}
\bibliographystyle{tmlr}

% =====================================================================
\appendix
\section{Proofs}
\label{app:proofs}

\paragraph{Proposition~\ref{prop:conv}.} $a_s$ is a function of $x_s$ alone and $c_s=g_s a_s = 0$ for $s\in S$, so no $c_v$ depends on $x_S$. Each $\tilde a_u$ is a function of the $c_v$ only; the bias of the convolution is a constant. $\delta_u$, $B_u$ and $C_u$ are linear in $\tilde a_u$, and $\Delta_u$ also involves the fixed $g_u$. The recursion for $h_u$ starts from $h_0=0$ and uses only $(\Delta, B, \tilde a)$, and $y_u$ uses $(C_u, h_u, \tilde a_u, g_u)$. The backward scan runs the same argument on the reversed sequence with the reversed, fixed gate. \hfill$\square$

\paragraph{Corollary~\ref{cor:logits}.} By Proposition~\ref{prop:conv}, $y_u$ does not depend on $x_S$ at any position. For an open token, $\tilde y_u = y_u\odot\mathrm{SiLU}(g_u z_u)$ with $z_u$ a function of $x_u$ only, so $\tilde y_u$ is free of $x_S$. For a closed token, $\tilde y_u$ depends on $x_u$ through $z_u$ unless $z$ is gated, in which case $\mathrm{SiLU}(0)=0$ gives $\tilde y_u=0$. The averaging of the two directions and the per-token LayerNorm map content-free vectors to content-free vectors. In Eq.~\ref{eq:pool}, a closed token contributes $w^j_u\tilde y^{(j)}_u$ to the numerator and $w^j_u$ to the denominator; both vanish under (i), and under (ii) the contribution is a constant. The concept logits are therefore invariant to $x_S$, and so is the task logit, which is a function of the concept probabilities only. \hfill$\square$

\paragraph{Proposition~\ref{prop:depth}.} With $z$ gated, $\tilde y_u = 0$ at closed positions, as above. Since $W_{\text{out}}$ has no bias, $W_{\text{out}}\tilde y_u = 0$, and with the residual gated, $g_u x_u = 0$. Hence $o_u = 0$ for $u\in S$ in each direction, whatever $x_u$ is. The next block therefore receives content-free input at those positions, and its LayerNorm maps it to a constant, even if its own gate re-opens them; Proposition~\ref{prop:conv} and Corollary~\ref{cor:logits} then apply to it with the content of the previous block's open positions as input. Without the $z$ and residual gates, $o_u$ at a closed position retains $x_u$, and a later block that re-opens the position passes it on. \hfill$\square$

\clearpage
\section{Implementation details}
\label{app:hparams}

{\small
\begin{longtable}{lp{0.64\textwidth}}
\caption{Settings of the final model and the baselines.}\\
\toprule
item & value \\
\midrule
\endfirsthead
\toprule
item & value \\
\midrule
\endhead
input & central axial slice of a $64{\times}64{\times}33$ cube at $0.75$\,mm, HU clipped to $[-1000,400]$, scaled to $[0,1]$ \\
tokens & $8{\times}8$ patches, 64 tokens, embedding dimension 192 \\
block & one bidirectional gated S6 block; inner width 384, state 16, $\Delta$-rank 12, causal conv width 4; mixers initialised from Vim-Tiny layer 0 \\
concept groups & 6 contiguous groups of 64 channels \\
gates & hard concrete, $\beta=2/3$, $\gamma=-0.1$, $\zeta=1.1$; evaluation rule of Eq.~\ref{eq:gate}; gate-logit bias initialised to 1 \\
Stage A & 150 epochs; concept BCE $+$ ramp $\times\,(3{\times}10^{-4}\cdot\text{hinge}(s+3) + 1.0\cdot(1-\text{recall}) + 5.0\cdot\mathrm{relu}(\bar g - 0.35))$; recall on masked images only; linear ramp over 20 epochs \\
Stage B & 100 epochs; frozen patch embedding, embedding norm and backbone gate; concept BCE $+$ ramp $\times\,(0.01\cdot\text{hinge on concept-gate logits} + 1.0\cdot(1-\text{Dice}))$; linear ramp over 20 epochs; masks for the same 50\% of training patients \\
optimiser & AdamW, lr $5{\times}10^{-4}$, weight decay $10^{-4}$, batch 16, 5-epoch linear warm-up, cosine to 2\%, gradient clipping 1.0 \\
label head & logistic regression ($C=1$, class-balanced) on predicted fold-training concept probabilities \\
concept thresholds & for binary concept AUROC: grid value in $\{2.5,3.0,3.5,4.0,4.5\}$ closest to a 50/50 split on fold-training data \\
augmentation & random $90^\circ$ rotations and flips \\
cross-validation & StratifiedGroupKFold (5 folds, grouped by patient), 3 seeds \\
baselines & SmallResNet (8 basic blocks, 16--128 channels) and SmallViT (width 192, depth 2, 3 heads, MLP 384), 150 epochs, same optimiser; unconstrained models use fold-level positive class weighting \\
ROAR-lite & top 16 tokens set to air before augmentation; unconstrained Mamba, same initialisation, 150 epochs; seed 0 \\
ROAR, shifted & as ROAR-lite, after a 24\,px shift of every crop in one of four directions (fixed per nodule, seed 2026), zero fill; shift applied before removal and augmentation \\
\bottomrule
\end{longtable}}

\section{Pre-declared tests}
\label{app:predeclared}

{\small
\begin{longtable}{lp{0.55\textwidth}ll}
\caption{All pre-declared tests, with the criterion as written before each run.}\\
\toprule
test & criterion, written before the run & result & section \\
\midrule
\endfirsthead
\toprule
test & criterion, written before the run & result & section \\
\midrule
\endhead
K1 & full, no-$\Delta$ and conv-only gating exact in every run & \pass & \ref{sec:res-ko} \\
K2 & no conv mask, and conv-only with $g^j$-only pooling, broken & \pass & \ref{sec:res-ko} \\
H3 & bottleneck cost $\le$ 1--2 points and below other backbones & partly & \ref{sec:res-acc} \\
Z1 & depth two, untrained blocks: conv-input, $z$ and residual gating exact without the $\Delta$ and $D$ gates (5 initialisations) & \pass & \ref{sec:res-ko} \\
Z2 & depth two: dropping either the $z$ or the residual gate leaks when block 2 re-opens tokens & \pass & \ref{sec:res-ko} \\
D2 & two blocks: (a) pooled task $\Delta\ge+0.01$, (b) concept AUROC $\Delta\ge+0.02$ & (a) \fail, (b) \pass & \ref{sec:res-acc} \\
M1, M2 & MambaLRP, hidden attention beat random ($\ge$12/15 runs) & \pass & \ref{sec:res-faith} \\
M3, M4 & gate beats MambaLRP, hidden attention ($\ge$12/15, $p<0.05$) & \fail & \ref{sec:res-faith} \\
G1 & corrected Grad-CAM beats random & \fail & \ref{sec:res-faith} \\
R1, R2 & ROAR: gate, hidden attention below random ($\ge$4/5 folds, DeLong $p<0.05$) & \pass & \ref{sec:res-faith} \\
R3 & ROAR: gate below centre prior (centred crops; the prior coincides with the nodule by construction) & \fail & \ref{sec:res-faith} \\
S1 & ROAR on 24\,px-shifted crops: gate below centre prior ($\ge$4/5 folds, DeLong $p<0.05$) & \pass & \ref{sec:res-faith} \\
S2 & ROAR on shifted crops: gate below random (same rule) & \pass & \ref{sec:res-faith} \\
P1, P2 & shifted crops: gate EPG above centre and intensity priors ($\ge$12/15) & \pass & \ref{sec:res-maps} \\
H2 & gates localise without mask supervision & \fail & \ref{sec:res-maps} \\
N1 & concept maps label-dependent (permuted below floor in $\ge$4/5 folds, gap $\ge0.05$) & \fail & \ref{sec:res-maps} \\
N2 & same, for the concept-specific contrast & \fail & \ref{sec:res-maps} \\
N3 & concept-specific contrast reproducible across seeds (floor $\ge0.3$) & \fail & \ref{sec:res-maps} \\
L1--L4 & partition lowers EXCESS and NEX, raw and PCA-64 (CI below 0, $\ge$4/5 folds; earlier recipe) & \pass & \ref{sec:res-leak} \\
F1 & final model: EXCESS below the oracle level (CI upper $<0$, $\ge$4/5 folds) & \fail & \ref{sec:res-leak} \\
3D & 3D crops match 2D & \fail & \ref{sec:discussion} \\
\bottomrule
\end{longtable}}

\section{ROAR per fold}
\label{app:roar}

\begin{table}[H]
\centering
\caption{Held-out AUROC of an unconstrained Mamba classifier retrained after removing the top 16 of 64 tokens under each ranking, nodule-centred crops (seed 0).}
\small
\begin{tabular}{lcccccc}
\toprule
tokens removed & fold 0 & fold 1 & fold 2 & fold 3 & fold 4 & pooled OOF \\
\midrule
none & 0.8525 & 0.9122 & 0.9076 & 0.8789 & 0.8407 & 0.8740 \\
random & 0.8609 & 0.8440 & 0.8681 & 0.7957 & 0.7957 & 0.8265 \\
gate & 0.7416 & 0.7253 & 0.7271 & 0.6946 & 0.7085 & 0.7206 \\
hidden attention & 0.7173 & 0.7101 & 0.6859 & 0.6693 & 0.7223 & 0.7017 \\
centre $4{\times}4$ & 0.6309 & 0.6347 & 0.6336 & 0.6733 & 0.6532 & 0.6466 \\
\bottomrule
\end{tabular}
\end{table}

\begin{table}[H]
\centering
\caption{The same experiment on crops shifted by 24\,px (three tokens) in one of four fixed directions per nodule; rankings recomputed on the shifted images; the reference (none) was retrained on shifted crops without removal (seed 0). Mean fraction of the nodule mask inside the removed set: gate $0.69$, hidden attention $0.69$, random $0.25$, centre $0.10$.}
\small
\begin{tabular}{lcccccc}
\toprule
tokens removed & fold 0 & fold 1 & fold 2 & fold 3 & fold 4 & pooled OOF \\
\midrule
none & 0.8407 & 0.8712 & 0.8577 & 0.8587 & 0.8426 & 0.8516 \\
random & 0.8502 & 0.8545 & 0.8337 & 0.7845 & 0.8188 & 0.8226 \\
gate & 0.6211 & 0.6387 & 0.6822 & 0.6621 & 0.7059 & 0.6655 \\
hidden attention & 0.6354 & 0.6530 & 0.6648 & 0.5793 & 0.6749 & 0.6426 \\
centre $4{\times}4$ & 0.8550 & 0.9055 & 0.8815 & 0.9161 & 0.8966 & 0.8859 \\
\bottomrule
\end{tabular}
\end{table}

\section{Knockout on a single validation split}
\label{app:ko-val}

\begin{table}[H]
\centering
\caption{Knockout on an earlier gated model trained without mask supervision; single validation split ($n=213$), one seed. AUROC differences at this sample size have a confidence interval of roughly $\pm0.06$ and are reported only for completeness.}
\small
\begin{tabular}{lccc}
\toprule
configuration & $\max|\Delta \hat c|$ (noise, SD 50) & guarantee & AUROC \\
\midrule
all paths gated & 0 & \ex & 0.8423 \\
conv-input mask removed & 4.25 & \br & 0.7467 \\
$\Delta$ gate removed & 0 & \ex & 0.7990 \\
$D$-skip gate removed & 0 & \ex & 0.8422 \\
$z$ gate removed & 0 & \ex & 0.8680 \\
residual gate removed & 0 & \ex & 0.8423 \\
conv input only & 0 & \ex & 0.8214 \\
all paths, $g^j$-only pooling & 0 & \ex & 0.7948 \\
conv input only, $g^j$-only pooling & 4.42 & \br & 0.8155 \\
no gating & 4.11 & \br & 0.7165 \\
\bottomrule
\end{tabular}
\end{table}

\end{document}